# Causality based Feature Fusion for Brain Neuro-Developmental Analysis

Peyman Hosseinzadeh Kassani, Li Xiao, Gemeng Zhang, Julia M. Stephen, Tony W. Wilson, Vince D. Calhoun, *Fellow, IEEE*, and Yu Ping Wang, *Senior Member, IEEE*

*Abstract*—Human brain development is a complex and dynamic process that is affected by several factors such as genetics, sex hormones, and environmental changes. A number of recent studies on brain development have examined functional connectivity (FC) defined by the temporal correlation between time series of different brain regions. We propose to add the directional flow of information during brain maturation. To do so, we extract effective connectivity (EC) through Granger causality (GC) for two different groups of subjects, i.e., children and young adults. The motivation is that the inclusion of causal interaction may further discriminate brain connections between two age groups and help to discover new connections between brain regions. The contributions of this study are threefold. First, there has been a lack of attention to EC-based feature extraction in the context of brain development. To this end, we propose a new kernel-based GC (KGC) method to learn nonlinearity of complex brain network, where a reduced Sine hyperbolic polynomial (RSP) neural network was used as our proposed learner. Second, we used causality values as the weight for the directional connectivity between brain regions. Our findings indicated that the strength of connections was significantly higher in young adults relative to children. In addition, our new EC-based feature outperformed FC-based analysis from Philadelphia neurocohort (PNC) study with better discrimination of the different age groups. Moreover, the fusion of these two sets of features (FC + EC) improved brain age prediction accuracy by more than 4%, indicating that they should be used together for brain development studies.

## I. INTRODUCTION

Human brain development is a prolonged process that is initiated from the third gestational week (GW) to late adolescence, and presumably to the entire lifespan [1]. A noteworthy period of life with significant level of brain development is from childhood to adulthood. For several decades, neuroscientists have been interested in how brain regions interact with each other and how brain connections change at this life stage. Several research endeavors have been made in the development of new methods to measure the brain transformation. Findings indicate that a few factors contribute to the brain maturation such as genetics, environment, and sex hormones [1, 2].

One of the most popular methods to probe brain functional connectivity is with functional magnetic resonance imaging (fMRI), which is a widely used non-invasive approach for investigating functional brain activity. Indeed, fMRI provides favorable insights into the ontogeny of functional brain regions [3].

Two popular types of brain connectivity measurement are functional connectivity (FC) and effective connectivity (EC). In FC, the temporal correlations between different functional regions are calculated from fMRI time series. FC also gives the average level of engagement of different brain regions [4]. In contrast to FC which does not assign directionality to the brain connections, EC describes which brain region leads (or is caused by) another one directly or indirectly [5, 6].

For FC-based approaches, Pearson correlation is often used to measure the temporal correlation between time series of different brain regions [7]. There are also several EC-based approaches to measure the flow of information distributed throughout the brain, among which Bayesian networks [8], Granger causality (GC) [9] and dynamic causal modelling (DCM) [10] are the most popular techniques. GC was introduced by Granger in 1969 [9]. GC uses temporal information to reveal causality influence of time series based on the multivariate autoregressive (MAR) model. Because of its simplicity, and easy implementation, GC has been extensively applied to neuroscience applications [5, 11-14]. Given two time series $\{X_t, Y_t\}_{t=1}^T$, if the inclusion of the history of $Y_t$ can improve the prediction of $X_{t+1}$, it is implied that the history of $Y_t$ contains unique information about $X_t$ [5]. In this sense, $Y_t$ Granger causes $X_t$, i.e., $Y_t \rightarrow X_t$, where $\rightarrow$ shows the direction of the connection. GC takes advantage of the least square error (LSE) minimization in the form of a simple linear model [15]. Hence, GC measures the *linear* causality values between brain regions.

However, with this definition, GC estimates could being

This work was partially supported by NIH (R01GM109068, R01MH104680, R01MH107354, P20GM103472, R01EB020407, 1R01EB006841) and NSF (#1539067).
(Corresponding author: Yu-Ping Wang)
P. H. Kassani, L. Xiao, G. Zhang, and Y.-P. Wang are with the Department of Biomedical Engineering, Tulane University, New Orleans, LA, 70118 USA (e-mails: peymanhk@tulane.edu , wyp@tulane.edu).

J. M. Stephen and V. D. Calhoun are with the Tri-institutional Center for Translational Research in Neuroimaging and Data Science (TReNDS), Georgia State University, Georgia Institute of Technology, Emory University, Atlanta, GA 30030, and also with the Department of Electrical and Computer Engineering, University of New Mexico, Albuquerque, NM 87131.
T. W. Wilson is with the Department of Neurological Sciences, University of Nebraska Medical Center, Omaha, NE 68198.

either severely biased or of high variance, leading to spurious results [16]. In most recent studies, the source of issues were reflected by the stationarity, linearity, noise, and sampling rates [13, 17]. Linear causation may fail to uncover more complicated causal relationships between different brain regions [14]. To consider the nonlinear complexity of brain network, in this paper we proposed a nonlinear kernel-based GC (KGC). KGC has been shown to be able to discover more causality patterns than the traditional GC [14]. In particular, we proposed a reduced Sine hyperbolic polynomial (RSP) neural network for this purpose. Multivariate polynomials (MP) and reduced polynomials (RP) [18] have been applied to several studies [19-21]. RSP is a new reduced polynomial to help learn the complex nature of data, by mapping the input vector into a nonlinear curve using sine hyperbolic.

The remainder of this paper is organized as follows: Section II includes the related works, motivation and contributions of this paper. Section III presents GC, and variants of polynomial classifiers as preliminaries. Section IV describes how we establish the proposed RSP to facilitate nonlinear causality analysis. Section V presents experimental results conducted on both synthetic and real data. Last section concludes this paper with some discussions.

## II. RELATED WORKS

Our GC-based fMRI analysis differs from previous FC-based brain neuro-development studies. Hence, we review related works on FC-based methods in this section, which will benefit to GC-based studies for examining brain neuro-development.

In [22], the age-related changes across development were measured by capturing correlated brain activity from resting state fMRI (rs-fMRI). First, principal component analysis was performed on the FC matrices across all subjects. Second, the reduced features as well as temporal features were extracted. Finally, three different regression models were utilized to make brain age prediction. According to [22], within-region connectivity was larger than between-region connectivity. At the edge-level of functional connectivity, the number of edges within default mode network (DMN) displayed linear decrement with age in older adults (subjects aged over 40) compared with younger adults (subjects aged under 40), which was consistent with previous studies [23, 24]. Linearly reduced functional connectivity in edges was also found within cingulo (CIN) network. Additional brain connections that showed linear increment with age were in between-region connections than within-region, especially between visual (VIS), somato-motor (SMN) and auditory (AUD) regions.

In the review [3], recent progress of the ontogeny of functional brain regions was summarized, which provided insights into the maturation of brain functional networks from childhood to adulthood. As reported in [3], the three most prominent regions from a developmental perspective are: (i) the frontoparietal central executive network (CEN) anchored in the dorsolateral prefrontal cortex and supra-marginal gyrus; (ii) the salience network (SN) anchored in the anterior insula and anterior cingulate cortex; and (iii) the DMN anchored in the posterior cingulate cortex, medial prefrontal cortex, medial temporal lobe, and angular gyrus [25, 26].

In the following, we describe some studies of EC-based brain analysis, as opposed to FC based analysis. In [12], the GC between cortical regions was measured. First, a least absolute shrinkage and selection operator (LASSO) was used to pre-select voxels. Next, a multivariate autoregressive (MAR) model was computed from the time series of the selected voxels. Finally, the Granger causality index (GCI) was calculated from the MAR model to represent directed inter-regional interactions. Results on both simulated and real data suggested that voxel-level signals better reflect the pattern of directed functional interactions between regions of interests (ROIs) than the voxel-averaged signals.

In [11], a new method based on GC was developed under the assumption that the history dependence varies smoothly. The main contribution of this study was that, the coefficients of the lagged history terms stem from smooth functions in MAR model. The history terms were modelled with the lower dimensional spline basis, which requires fewer parameters than the standard approach. This procedure allows accurate estimation of brain dynamics and functional networks in both simulations and real analysis of brain voltage activity recorded from a patient with epilepsy. The proposed GC method has more statistical power than the original GC for networks with extended and smooth history dependencies.

From the above review, it is realized that several studies discuss FC-based brain age development using temporal correlation, but there is a lack of EC-based approaches. Our goal is therefore to discover the nonlinear causal interactions within and between brain regions. This motivated us to propose RSP for EC-based feature extraction.

The contributions of this study are outlined as follows:
1) To extract EC-based features, reduced Sine hyperbolic polynomial (RSP) was used to learn the nonlinear nature of fMRI data. We showed the directional connectivity constructed after Power264 parcellation [27]. We found that three brain networks, i.e., DMN, visual attention (VIS), and ventral attention (VENT) are often leading other brain networks' activity. On the other hand, four brain regions, i.e., DMN, salience (SAL), auditory (AUD), and VENT, are often caused by other brain regions. Roughly 25% of the whole brain directional connections are intra-connections within DMN. For between-region connectivity, the pair of DMN and VIS, and the pair of DMN and VENT have most directional connections. We found memory and dorsal attention regions have the least number of connections for brain maturation. All the experimental results were supported by the statistical t-test.
2) Interestingly, for up to 95% of brain directional connections, young adults (over 18) have stronger connections between and within brain regions than children (below 13) when causality values are defined as the weights of edges between brain regions. This difference is more significant in DMN for both within and between region connections.
3) For our simulated data, all competing methods, to our knowledge, correct causality for linear relationship, while

for the case of nonlinear relationship, the traditional GC had poor performance. This implies the necessity of using kernel based methods for causal discovery. For Philadelphia Neurodevelopmental Cohort (PNC) data analysis, our EC-based feature extraction with RSP outperformed FC-based one (with 1% gap in the classification accuracy), resulting in better separation of two age groups. This emphasizes the importance of directional flow in assessing brain connections.

4) The fusion of two sets of features (FC + EC) further improved brain age prediction accuracy by more than 4% accuracy. The large gap in differences between EC-based and fused-based features, that is 4%, motivated us to compare the differences between EC-based and fusion feature-based brain connections. Findings supported by the t-test statistics showed us new causations between brain regions and also misleading causations found by merely EC-based ones. After removing those new features from the fusion-based features, we classified the age groups again and this time, the accuracy dropped by nearly 3%. This proved the importance of using fusion-based features for the task of brain maturation study.

5) The experiments conducted on both simulated data and real-world PNC data further validated the reliability of the model used in this study.

## III. PRELIMINARIES

### A. Granger causality (GC) analysis in brain fMRI

Granger causality test is based on a linear MAR model to discover underlying linear causal relations. Let $B_i(t)$ express the BOLD response of ROI $i$ at time point $t$. With a linear combination of $p$ previous blood oxygen level dependent (BOLD) responses, $B_i(t-1), \ldots, B_i(t-p)$, one can predict the value of $B_i(t)$ as follows:

$$B_i(t) = \sum_{k=1}^{p} a_{ik} B_i(t-k) + \varepsilon_i(t) \quad (1)$$

where $\varepsilon_i(t)$ is the error term and obeys a Gaussian distribution with mean zero and variance $\sigma_i^2$. The error term $\varepsilon_i(t)$ is used as the measurement of the accuracy of prediction. For a bivariate MAR model, assume that the response in ROI $j$ causes activation in ROI $i$. If so, the addition of the prior BOLD response from region $j$, i.e., $B_j(t)$, to Eq. (1), should improve the prediction power of $B_i(t)$. The improvement in the prediction of $B_i(t)$ simply means that the error variance $\sigma_i^2$ should decrease [28]. The augmented linear combination of both ROIs $i$ and $j$ is as follows,

$$B_i(t) = \sum_{k=1}^{p} \left[ a_{jk} B_j(t-k) + a_{ik} B_i(t-k) \right] \quad (2)$$
$$+ \varepsilon_{i|ij}(t)$$

where $\varepsilon_{i|ij}(t)$ denotes the error term when $B_j(t)$ is added to the model and $\varepsilon_{i|ij}(t)$ is also drawn from a Gaussian distribution with mean zero and variance $\sigma_{i|ij}^2$. The purpose of using Equations 1 and 2 is to determine if activation in ROI $i$'s signal causes activation in ROI $j$'s signal. Geweke in 1982 [29] suggested to use the following mathematical expression to measure causality,

$$F_{j \to i} = \ln \left( \frac{\sigma_i^2}{\sigma_{i|ij}^2} \right) \quad (3)$$

where the notion $F_{j \to i}$ is termed as Granger causality index (GCI). It is always true that $\sigma_i^2$ is not less than $\sigma_{i|ij}^2$ since Eq. (2) has more parameters than Eq. (1). Hence, $F_{j \to i} > 0$. In addition, if region $j$ has no influence on region $i$, then adding prior observations of $B_j(t)$ will not improve the prediction of current value of $B_i(t)$ and thus, $F_{j \to i} = 0$. Putting all together, $F_{j \to i} \geq 0$. The larger the value of $F_{j \to i}$ is, the stronger the causation of region $j$ over region $i$ is.

Suppose the lag order $p$, is an arbitrary value and fMRI has also $n$ discretized sampled responses. In matrix form, the model with a single ROI $i$ can be written as,

$$\begin{bmatrix} B_i(n) \\ B_i(n-1) \\ \vdots \\ B_i(p+1) \end{bmatrix} =$$

$$\begin{bmatrix} B_i(n-1) & \cdots & B_i(n-p) \\ \vdots & \ddots & \vdots \\ B_i(p) & \cdots & B_i(1) \end{bmatrix} \times \begin{bmatrix} a_{i1} \\ a_{i2} \\ \vdots \\ a_{ip} \end{bmatrix} + \begin{bmatrix} \varepsilon_i(n) \\ \varepsilon_i(n-1) \\ \vdots \\ \varepsilon_i(p+1) \end{bmatrix} \quad (4)$$

After the inclusion of ROI $j$ into Eq. (4) as an augmentation to the prediction of ROI $i$,

$$\begin{bmatrix} B_i(n) \\ B_i(n-1) \\ \vdots \\ B_i(p+1) \end{bmatrix} =$$

$$\begin{bmatrix} B_i(n-1) & B_j(n-1) & \cdots & B_i(n-p) & B_j(n-p) \\ \vdots & & \ddots & \vdots & \\ B_i(p) & B_j(p) & \cdots & B_i(1) & B_j(1) \end{bmatrix} \times \begin{bmatrix} a_{i1} \\ a_{j1} \\ \vdots \\ a_{ip} \\ a_{jp} \end{bmatrix} \quad (5)$$

$$+ \begin{bmatrix} \varepsilon_{i|ij}(n) \\ \varepsilon_{i|ij}(n-1) \\ \vdots \\ \varepsilon_{i|ij}(p+1) \end{bmatrix}$$

In compact form, it can be rewritten as

$$\underline{y} = \mathbf{X}\underline{\boldsymbol{\beta}} + \underline{\boldsymbol{\varepsilon}} \quad (6)$$

where $\underline{y}$ in Eq. (6) is equal to the left hand side of Eq. (5) and $\mathbf{X}$, $\underline{\boldsymbol{\beta}}$, and $\underline{\boldsymbol{\varepsilon}}$ are corresponding to fMRI BOLD responses $B$, coefficients $a$ and error variance $\varepsilon$, respectively. The unknown coefficients $a_{ik}$ or $\underline{\boldsymbol{\beta}}$, can be estimated through least square minimization as follows:

$$\widehat{\underline{\beta}} = (\mathbf{X}^T\mathbf{X})^{-1}\mathbf{X}^T\underline{y} \quad (7)$$

*B. Multivariate Polynomial (MP)*

Polynomial models can be used to approximate any complex nonlinear relationship in data. Indeed, the polynomial models are the Taylor series expansion of a function [30]. Multivariate polynomial neural network uses the polynomial of an order for kernel-based learning. The size of the feature depends on the order of the polynomial. This feature expansion helps to learn the nonlinear nature of data [18]. Let $\mathbf{x} \in \mathcal{R}^d$ denote the row vector of the matrix $\mathbf{X} \in \mathcal{R}^{N \times d}$, where $d$ is the input column vector before using the polynomial-based projection. With the $r$-th order of polynomial, $\mathbf{x} \in \mathcal{R}^d$ is mapped to an expanded vector $\mathbf{q} \in \mathcal{R}^D$, where $D = \binom{d+r}{r}$ and (.) is the combinatorial counting. With this definition, the matrix $\mathbf{X} \in \mathcal{R}^{N \times d}$ is mapped or projected to $\mathbf{Q} \in \mathcal{R}^{N \times D}$. The rows of $\mathbf{Q} \in \mathcal{R}^{N \times D}$ are denoted by vectors $\mathbf{q} \in \mathcal{R}^D$. To predict $\mathbf{y} \in \mathcal{R}^N$, there should be a weight coefficient vector $\mathbf{w} \in \mathcal{R}^D$ with the following relation to $\mathbf{q} \in \mathcal{R}^D$,

$$z(\mathbf{w}, \mathbf{x}) = \sum_{k=0}^{D-1} w_k q_k(\mathbf{x}) = \mathbf{q}(\mathbf{x})^T \mathbf{w} \quad (8)$$

where $z(\mathbf{w}, \mathbf{x})$ is the estimated output and $\mathbf{q}(\mathbf{x})$ is the projected vector of $\mathbf{x}$. Using the sum of square errors (SSE) as an objective function, the distance between $z(\mathbf{w}, \mathbf{x})$ to the real output $y$ is minimized for vector $\mathbf{q}(\mathbf{x})$. In matrix form, the SSE objective function can be written as,

$$obj(\mathbf{w}) = \tfrac{1}{2}\|\mathbf{y} - \mathbf{Qw}\|_2^2 \quad (9)$$

Differentiating $obj(\mathbf{w})$ in Eq. (9) w.r.t $\mathbf{w}$, the optimal weight vector $\mathbf{w} = (\mathbf{Q}^T\mathbf{Q})^{-1}\mathbf{Q}^T\mathbf{y}$ is obtained.

### IV. PROPOSED REDUCED SINE HYPERBOLIC POLYNOMIAL (RSP) FOR CAUSALITY LEARNING

To avoid undesirable growth of features generated by the polynomial method, its reduced variant, called RP [18], is employed. RP takes advantage of the mean value theorem and seeks for effective features that have more influence on the learning procedure. Suppose $f(\mathbf{\gamma}) = (\gamma_{j1}x_1 + \gamma_{j2}x_2 + \cdots + \gamma_{jd}x_d)^j, j = 2, \ldots, r$, and given two arbitrary points $\mathbf{\gamma}$ and $\mathbf{\gamma}_1$, the polynomial function is written as,

$$f(\mathbf{\gamma}) = f(\mathbf{\gamma}_1) + (\mathbf{\gamma} - \mathbf{\gamma}_1)^T \nabla f(\bar{\mathbf{\gamma}}) \quad (10)$$

where $\bar{\mathbf{\gamma}} = (1 - \vartheta)\mathbf{\gamma}_1 + \vartheta\mathbf{\gamma}$, with $0 \le \vartheta \le 1$. By discarding $\mathbf{\gamma}_1$ and the polynomial terms between $f(\mathbf{\gamma}_1)$ and $\nabla f(\bar{\mathbf{\gamma}})$, a less complicated polynomial function $\hat{f}(\mathbf{\gamma})$ is obtained as follows [18]:

$$\hat{f}_{RSP'}(\mathbf{\gamma}, \sinh(\mathbf{x}))$$

$$= \gamma_0 + \sum_{j=1}^{d} \gamma_j \sinh(x_j)$$

$$+ \sum_{j=1}^{r} \gamma_{l+j}(\sinh(x_1) + \sinh(x_2) \quad (11)$$

$$+ \cdots + \sinh(x_d))^j$$

$$+ \sum_{j=2}^{r} \left(\mathbf{\gamma}_j^T \cdot \sinh(\mathbf{x})\right)(\sinh(x_1)$$

$$+ \sinh(x_2) + \cdots + \sinh(x_d))^{j-1}$$

Note that we replaced the input feature $\mathbf{x}$ with the sine hyperbolic function which is defined as $\sinh(\mathbf{x}) = \frac{e^{\mathbf{x}} - e^{-\mathbf{x}}}{2}$ and hence it is called reduced Sine hyperbolic (RSP).

The new polynomial $RSP'$ has $D = 1 + r(d - 1)$ terms. To improve the learning ability, the authors in [18] added more exponential terms to Eq. (11) as follows:

$$\hat{f}_{RSP}(\mathbf{\gamma}, \sinh(\mathbf{x}))$$

$$= \gamma_0 + \sum_{k=1}^{r}\sum_{j=1}^{d} \gamma_j \sinh(x_j^k)$$

$$+ \sum_{j=1}^{r} \gamma_{rd+j}(\sinh(x_1) \quad (12)$$

$$+ \sinh(x_2) + \cdots + \sinh(x_d))^j$$

$$+ \sum_{j=2}^{r} \left(\mathbf{\gamma}_j^T \cdot \mathbf{x}\right)(\sinh(x_1)$$

$$+ \sinh(x_2) + \ldots + \sinh(x_d))^{j-1}$$

With this setting, RSP has only $D = 1 + r + d(2r - 1)$ parameters, where $d$ is the number of columns or features in the input space before the polynomial-based projection. RSP suffers less from the overfitting problem as $D$ and $r$ have linear relationship, while in multivariate polynomial $D$ and $r$ have exponential relationship.

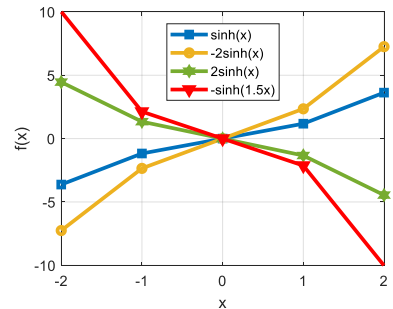

Fig. 1. Reduced Sine hyperbolic polynomial x with different parameter values can generate complicated curves for learning nonlinearity in the brain network

Note that each variable $\mathbf{x}$ in the input space is projected to a new space using RSP. In fact, a straight line is mapping to a curve through the sine hyperbolic function; as a result, it can learn more complicated data as displayed by Fig. 1. In the experiments, we use η × Sinh(σx) where η and σ are two adjustable parameters.

Fig. 2 shows a schematic illustration of the proposed framework in this study. The raw rs-fMRI images are first preprocessed and then are divided into two categories, namely subjects with age below 13 and subjects with age above 18.

The reason to divide the subjects into two age groups is that, a major transition in brain maturation takes place during the second decade of human life [31]. During this period, the adolescents' brain encounters changes in physiology, emotional and cognitive skills as well as in reasoning and decision making [31]. As an example, Fair et al. [32], used ROI-based analyses to examine differential connectivity of two age groups, i.e., 7- to 9-year-old children compared to 21- to 31-year-old adults. Therefore, we can better discern the differences in brain connectivity between groups. The Granger causality (GC) matrix is obtained through polynomial based method (e.g., proposed RSP) for each subject in each category. Doing this for $N$ subjects, we obtain $N$ causality matrices. Vectorizing each matrix yields a new matrix with size $N \times F$. Next, the matrix of size $N \times F$ is divided into training and testing sets for 10-fold cross validation. After that, a support vector machine (SVM) with linear kernel (SVM-Lin) is used to learn the differences between two groups in terms of classification accuracy. It is worth mentioning that RSP has tuning parameters η, σ, which lead to different GC matrices. In real data analysis, the model selection is based on the cross-validation. That is, the best model parameters are those that can give the highest classification accuracy. The corresponding GC matrix is computed based on this model.

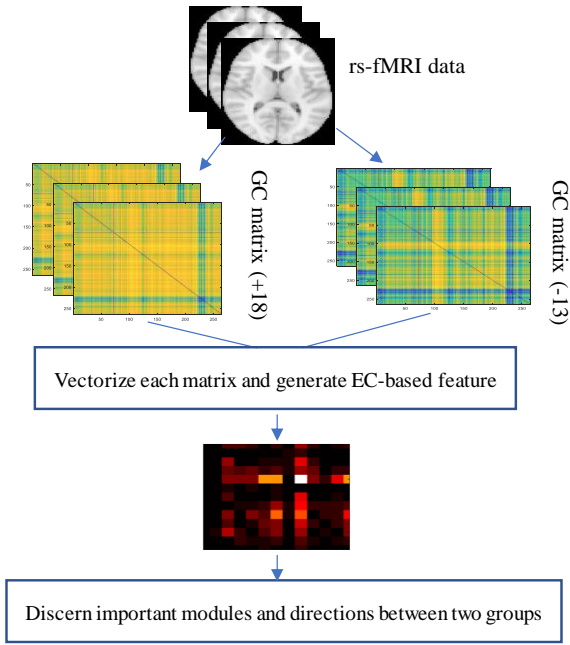

Fig. 2. The flowchart of the proposed method for brain development analysis.

## V. Experiments

### A. Datasets and Setup

The rs-fMRI data used in this study are from the Philadelphia Neurodevelopmental Cohort [33], which is a collaborative research project between the Brain Behavior Laboratory at the University of Pennsylvania and the Center for Applied Genomics at the Children's Hospital of Philadelphia [30]. The data are available in the dbGaP database. There are nearly 1000 adolescents within the range from 8 to 21 years. The voxel matrix for each subject is of 64×64×46, and with voxel size = 3mm$^3$. The whole brain scanning session takes minutes. Since the purpose of this study is brain development, we are interested in individuals with ages lower than 13 and higher than 18 for our experiment. Table I tabulates some characteristics of the two age groups.

TABLE I
CHARACTERISTICS OF THE SUBJECTS IN THIS STUDY. SD: STANDARD DEVIATION

| Group | Age (Mean ± Std) | Gender (M/F) |
|---|---|---|
| Child (-13) | 9.84 ± 0.67 | 62/69 |
| Young Adult (+18) | 19.35 ± 1 | 73/114 |

In total, 264 ROIs (containing 21,384 voxels) were extracted based on the Power parcellation. The reduction in dimensionality from voxels to ROIs was based on a sphere radius parameter of 5 mm. Standard brain imaging processing techniques including motion correction, spatial normalization to standard MNI space (spatial resolution $2 \times 2 \times 2$mm) and spatial and temporal smoothing with a 3mm FWHM Gaussian kernel were performed on fMRI data through SPM12 [27].

Some notes for the experimental setup include:
1. For the sake of comparison between dense and sparse causality maps, the performance of using the polynomial based models, MP, RP, and RSP is compared.
2. The lag order $p$ is chosen by Bayesian Information Criteria (BIC) during the learning procedure and the optimal $p$ is mostly 1.
3. For polynomial based learners, the polynomial order $r$ is in the range from 1 to 5 and the best value of $r$ is used to predict the testing labels. To find the best value for $r$, the learners are run 10 times and the mean of GCI is taken into consideration.
4. For the proposed RSP, the parameters **η** and **σ** are both in the range [0.1, 0.2, … , 1] with step size of 0.10.
5. For FC-based feature extraction, the correlation between brain regions is calculated with Pearson correlation.
6. The proposed model is implemented with MATLAB on a high-performance computer, called Cypress (Cypress: High Performance Computing System, https://crsc.tulane.edu/) with 8 nodes of CPU, and 64 GB RAM.

### B. Simulation study

Similar to the study in [14], we first evaluate all learners on a synthetic data with both linear and non-linear causalities.

#### 1) Linear causality model data

We consider three time series obtained from the autoregressive processes with three variables ($X_1, X_2, X_3$) with linear relationships as follows [14]:

$$X_1(t) = 0.441 X_1(t-1) + 0.02\tau_1(t);$$
$$X_2(t) = 0.8 X_1(t-1) + 0.02\tau_2(t); \quad (13)$$
$$X_3(t) = -0.7 X_1(t-1) + 0.02\tau_3(t);$$

where $\tau$'s are the unit variance noise drawn from Gaussian distribution. The causal relationships for this linear simulated data are therefore specified to be $1 \to 2$ and $1 \to 3$. The total number of time points is set to $M = 1000$ and the number of iterations is set to 50.

#### 2) Non-linear causality model data

Three time series are obtained from the autoregressive processes with three variables ($X_1, X_2, X_3$) with non-linear relationships defined as follows:

$$\begin{aligned}
X_1(t) &= (1-e)(1 - aX_1^2(t-1)) \\
&\quad + e(1 - aX_2^2(t-1)) + s\tau_1(t); \\
X_2(t) &= 1 - aX_2^2(t-1) + s\tau_2(t); \quad (14) \\
X_3(t) &= (1-e)(1 - aX_3^2(t-1)) \\
&\quad + e(1 - aX_1^2(t-1)) + s\tau_3(t);
\end{aligned}$$

where $a = 1.8$, $s = 0.02$, $e = 0.2$ and the $\tau$'s are the unit variance noise drawn from a normal distribution. The causal relationships for this non-linear simulated data are $2 \to 1$ and $1 \to 3$. $M = 1000$ and 50 runs of the models are conducted.

#### 3) Results on simulated data

The proposed RSP is compared to traditional MP and RP, and linear algorithms for discovering causality. To generate a linear model, we use MP with order 1, as the linear model in order to demonstrate how non-linearity is important for the task of causality discovery. Fig. 3 displays this comparison on both linear and non-linear simulated data.

As can be seen from Fig. 3, for the case of linear relationship within the simulated data, all methods give similar GCI values and can successfully detect causal relationships. Note that RSPF refers to RSP with fixed parameter values (i.e., all parameter values are equal to 1).

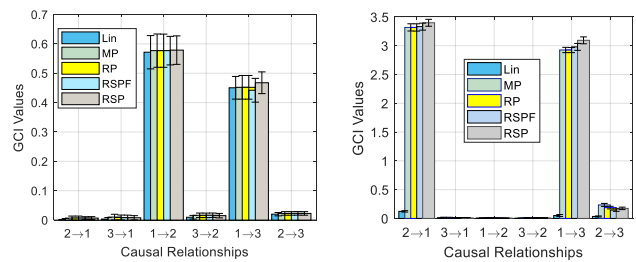

Fig. 3. The comparison of the four GC models, left: linear and right: non-linear simulated data.

For the simulated nonlinear relationship, orthogonal least square learner with linear kernel (OLS-Lin) completely fails to detect causal relationships. This means that a linear model is unable to detect non-linear causalities. But other counterparts can clearly detect such causal relationships. Our proposed RSP is slightly better than other counterparts. Due to the nonlinear nature of brain connected regions, we discard OLS-Lin from our modelling pool for real data analysis in the next subsection.

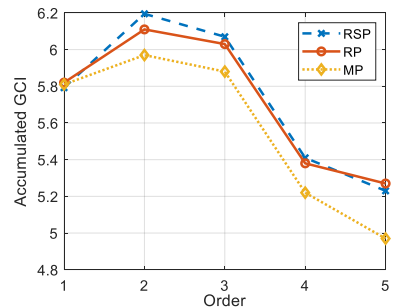

Fig. 4. The effects of polynomial order with linear data; a binomial learner (when the order of polynomial is equal to 2) is optimal

In Fig. 4, the GCI value is the sum of all values of the 3×3 Granger causality matrix and is hence called accumulated GCI. It should be noted that the diagonal elements of the GC matrix are zeros.

### C. fMRI real data experiments

#### 1) Parameter selection

For real data, we do not have the prior knowledge or ground truth for causality analysis between brain regions for two age groups. Hence, a binary classifier is performed to choose the optimal parameters for every learner, namely MP, RP, and the proposed RSP. The two classes are subjects with age above 18 (positive class) and under 13 (negative class).

First, for each subject, the causality matrix ($264 \times 264$) is vectorized into a row vector with length of 69696. Second, for all subjects, the size of data matrix is $318 \times 69696$. Finally, a support vector machine classifier with linear kernel (SVM-Lin) is performed to classify subjects into either young adults or children groups. We run each model 100 times with 10-fold cross-validation. For each model, 100 test accuracies are obtained. The average of accuracy is reported as the final result. The selection of optimal parameters is based on the best accuracy given by SVM-Lin. The causality matrix corresponding to the optimal parameters is then chosen for brain neurodevelopment analysis.

Examining the results in Table II, among EC based features, RSP performs slightly better than MP and RP. The classification accuracy of RSP based EC feature is one percent better than the correlation based FC feature. We further concatenate FC and EC based features. The combination of FC feature and EC-based feature gains 0.929 accuracy, much better than either using a single EC with accuracy of 0.883 or FC with accuracy of 0.877. So, we claim that the fusion of EC and FC based features gives better accuracy than either one for brain age prediction. The rationale behind this better accuracy is the following. Intuitively, the FC based features only give us information about the statistical dependency of brain ROIs.

When we add EC based features to FC, the directions of the connections will provide more information by specifying how a brain region affects another in a directional flow. Hence, a more discriminative feature is generated.

TABLE II
THE PERFORMANCE OF CLASSIFIERS ON TWO BRAIN AGE GROUPS

| Model Name | Acc |
|---|---|
| **Effective Connectivity (EC)** | |
| MP | 0.873 ± 0.060 |
| RP | 0.874 ± 0.057 |
| RSP | 0.883 ± 0.052 |
| **Functional Connectivity (FC)** | |
| Corr | 0.877 ± 0.0501 |
| **FC + EC** | |
| Corr+MP | 0.915 ± 0.048 |
| Corr+RP | 0.921 ± 0.049 |
| Corr+RSP | **0.929** ± 0.041 |

*2) Results on resting state fMRI from PNC*

Since the proposed RSP is the winner among all learners, it is used for the subsequent analysis of brain maturation. To do so, a t-test is employed to show differences between the two age groups.

We removed bi-directional connectivity (i.e., simultaneous $x \rightarrow y$ and $y \rightarrow x$) and the non-zero uni-directional connections are used as the final features for the analysis.

We divided the 264 brain regions via POWER parcellation into 13 functional subnetworks. These regional connectivity are somatomotor hand (SMA/H), somatomotor mouth (SMA/M), CIN, auditory (AUD), default mode network (DMN), membership retrieval (MEM), visual (VIS), frontal parietal network (FPN), salience (SAL), subcortical (SBC), ventral attention (VNT), dorsal attention (DRS), and cerebellar (CRB).

The outcome of the t-test is displayed in Fig. 5. DMN brain region has the maximum number of connections for both within- and between-region connections. This means that DMN more frequently causes other brain regions while it is also often led by other regions.

The second and third brain regions that are causal to other regions are VIS and VNT respectively. We also see that DMN, SAL, AUD, and VNT are frequently led by other regions respectively. Additionally, the brain regions that have the least connections and causations are MEM and DRS.

Fig. 6 displays the circular graph for good visualization of the strength of connections between the two most important brain regions DMN and VIS. Among 560 remaining uni-directional brain connections, 68 of them are between DMN and VIS.

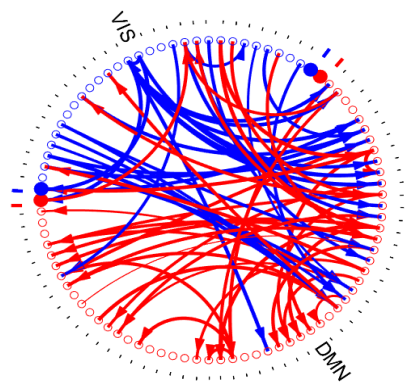

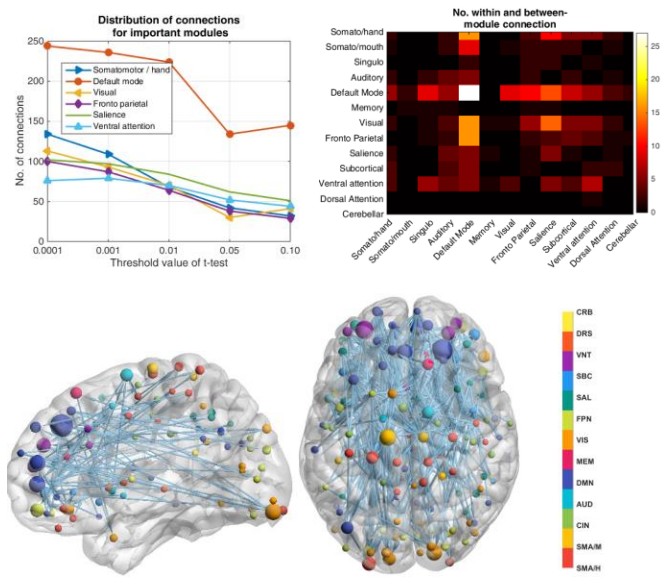

Fig. 5. The brain regions associated with the EC-based connections show a large difference between the low and high age groups. In the first row, the left figure shows the number of connections through different threshold values of t-test and the right figure displays the number of within- and between-region connections. The second row displays medial and axial view in anatomical space for only those brain ROIs that are statistically different. The node colors indicate region membership.

Fig. 6. Directional circular graph between regions in the two most important brain networks, DMN and VIS

From Fig. 6, DMN has much more within-region connections than VIS. DMN is less causal for VIS while the number of directional connections of VIS that are causal for DMN is higher than its within-region connections.

Fig. 7 shows $GCI_{high} - GCI_{Low}$ the difference of the GCI values between the two age groups. Fig. 7 displays how the strength of brain region connection increases as the brain develops with age. This is in line with studies considering FC-based brain maturation [34-36]. For example, the study in [35] found that functional connectivity weakens for short-distances between brain regions in the young brain while it is stronger for long-distance functional connectivity in the older brain. A study described in [36] performs a comprehensive analysis for the comparison between two age groups. The correlation matrices generated from child and adult rs-fMRI data reveal that nodes within the DMN are sparsely connected in children, and strongly connected in adults. Similarly, in our study, the difference of causality values between two age groups is significant. From 560 remaining connections which are statistically different between two age groups, we only found less than 10 connections in which the $GCI_{low}$ are larger than the $GCI_{high}$. This means as the brain develops with age, stronger directional connections form in the brain.

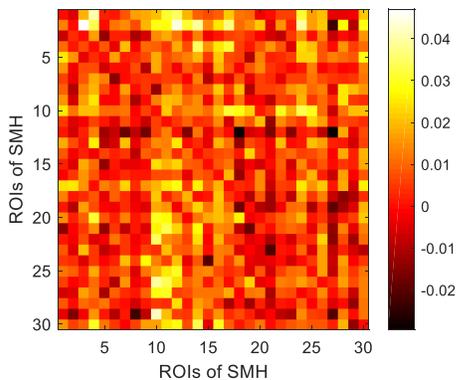

Fig. 7. Difference in strength of the GCI values between groups of young adults and children for somotomotor hand (SMH) subnetwork.

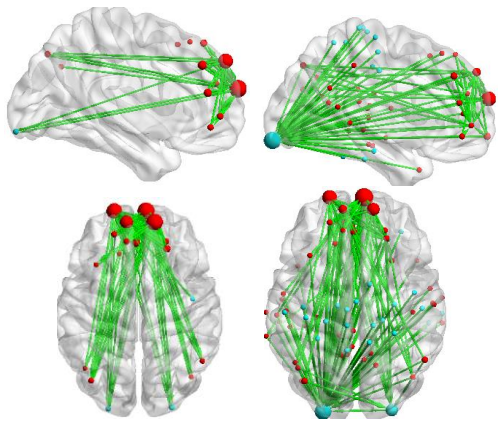

Fig. 8. Different brain anatomies and directional connections between DMN (red) and VIS (blue). Left column: Young adults and Right column: Children

Fig. 8 displays brain anatomies from different views, where DMN leads VIS (DMN → VIS). Obviously, the majority of connections belong to the within-region connections in DMN region[37], similar to the circular graph in Fig. 6. The size of a node indicates the number of connections coming to the node or going out of that node, which is defined as node degree node.

### D. Fusion (EC + FC)-based features versus EC- based features

Following Table II, since the accuracy obtained by fusion-based features is 4 percent better than that of both EC- and FC-based features, we are interested in producing new sets of EC-based features in terms of the fusion of EC and FC.

To do so, we compare the causality matrices that were obtained by the *t*-test for both EC and FC features. Our main assumption is, those highly correlated features tend to give directional connections between two different brain ROIs. Hence, two t-tests at a significance level of 0.01 with false discovery rate (FDR) of 0.05 are performed on both FC- and EC-based features and their intersections/common features are taken into consideration. Applying *t*-test, two binary matrices with size 264 × 264 are obtained for both FC- and EC-based features. Then, an element-wise comparison is performed on both binary matrices. Only the elements that have '1', i.e., statistical difference between two brain ROIs, implying both high correlation and high causation, were picked up to filter out the EC-based features. There were EC-based 6338 connections and after filtering by the correlation matrix, only 1490 connections (EC + FC) are left for display in Fig. 9.

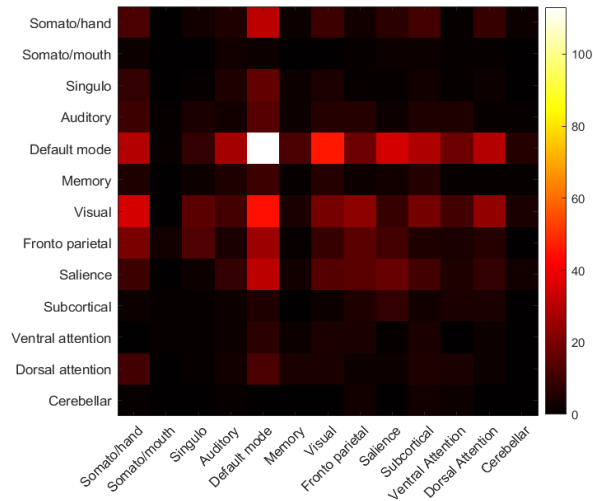

Fig. 9. Difference in strength of the GCI values - group of subjects with high age minus subjects with low age.

All the directional connections in the causality matrix obtained by EC-based features with low values have been deleted. Again, DMN and VIS are among brain regions with the two largest number of connections. However, the brain region Somato/mouth lost several directional connections, implying low correlation of its connections.

In Table III we list 10 significantly connected areas between DMN and SMH. These connections were derived from both FC- and EC-based features. They brought information for the district, position and MNI coordinates of each directional connection.

TABLE III
SIGNIFICANT AREAS OF SMH ARE CAUSED BY DMN OR CAUSING DMN

| Significantly connected area | District (Lobe) | Position (Cerebrum) | MNI Coordinates | | |
|---|---|---|---|---|---|
| | | | X | Y | Z |
| Regions influenced by the seed DMN (X->Y) | Limbic | L | -14 | -18 | 40 |
| | Parietal | L | -23 | -30 | 72 |
| | Parietal | R | 47 | -30 | 49 |
| | Parietal | R | 13 | -33 | 75 |
| | Parietal | L | -40 | -19 | 54 |
| Regions influencing the seed DMN (Y->X) | Limbic | L | -13 | -40 | 1 |
| | Temporal | R | 46 | 16 | -30 |
| | Parietal | R | 6 | -59 | 35 |
| | Limbic | L | -2 | -37 | 44 |
| | Frontal | L | -10 | 39 | 52 |

L: Left, R: Right

In **supplementary materials**, we include more brain regions and the connections between them in terms of our new finding.

We are also interested in detecting highly connected nodes, i.e., the hubs. Similar to [38], we also defined hubs that are nodes with degrees having at least two standard deviations higher than the mean. In this way, we identified 10 hubs as in Table IV. Most of these hub ROIs are from the frontal lobe, Parietal lobe, Occipital lobe, and limbic lobe, and sub lobar respectively with much larger connections or degrees than the average. This finding is in line with the study in [36] that also

TABLE IV
THE HUB ROIS IN ROI NETWORK. DG REPRESENTS DEGREE

| ROI index | ROI Name | DG |
|---|---|---|
| 10 | Parietal Lobe-Postcentral Gyrus(R)-SMH | 154 |
| 181 | Frontal Lobe-Middle Frontal Gyrus(L)-FPT | 143 |
| 220 | Frontal Lobe-Inferior Frontal Gyrus(R)-VNT | 139 |
| 185 | Frontal Lobe-Sub Gyral(R)-SAL | 131 |
| 232 | Frontal Lobe-Middle Frontal Gyrus(R)-DRS | 126 |
| 4 | Limbic Lobe-Cingulate Gyrus(R)-SMH | 126 |
| 66 | Limbic Lobe-Parahippocampa Gyrus(L)-DMN | 118 |
| 148 | Occipital Lobe-Lingual Gyrus(R)-VIS | 116 |
| 206 | Sub Lobar-Lentiform Nucleus(L)-SBC | 113 |
| 137 | Occipital Lobe- Middle Occipital Gyrus (L)-VIS | 112 |

L: Left, R: Right

compared children with young adults, but only with FC features.

To prove the importance of new findings, i.e., whether highly correlated features give us accurate causality, we remove all 1490 features from the pool of EC-based features. Specifically, every time by the removal of 10% features, we run SVM-linear with 10-fold cross validation. We also randomly remove 1490 EC-based features every time by the removal of 10% features. This helps us to verify how the accuracy of separating two age groups changes and how effective the highly correlated and causal features are. In fact, we want to show that *correlation complements causation*. Fig. 10 shows the changes in classification accuracy.

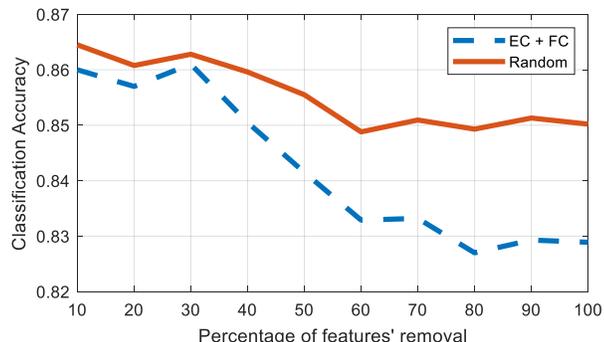

Fig. 10. Percentage of features' removal versus classification accuracy with random feature pruning and fusion based feature pruning. The drop in accuracy emphasizes the importance of fused based (EC+FC) features.

As can be seen from Fig. 10, the drop in accuracy is roughly 2% between the removal of random features and (EC + FC)-based ones. This implies that the correlation information helps for the detection of directional connections. Additionally, in Table II, we observed that combining *correlation* with *directional* information by EC-based features helps to improve the classification accuracy between the two age groups by 4%.

Studies have shown that adults have weak short-range functional connectivity but strong long-range functional connectivity, known as functional integration. In children, it is reversed, i.e., strong short-range functional connectivity but weak long-range functional connectivity, known as functional segregation [36]. To verify this in our experiment, we treat brain networks as a graph, and use quantifying measures such as local efficiency and global efficiency as tested by other studies [39]. By definition, global efficiency is the average of inverse shortest path length whereas local efficiency is the global efficiency, which is only computed on the neighborhood of the each node of graph and is related to the clustering coefficient [39].

To do so, we use Brain Connectivity Toolbox (https://sites.google.com/site/bctnet/Home) [39]. We deal with a weighted graph, i.e., the GCI value is treated as the weight of each edge of the graph. We use the global efficiency, $E_{Glob}$ and local efficiency, $E_{Loc}$, for a weighted directional graph. Each subject has a causality matrix which is treated as a weighted directional graph. The $E_{Glob}$ is computed for each graph and for each age group. Doing so, we obtain two vectors of $E_{Glob}$, one for each age group. A t-test with a significance level of 0.01 is performed to test if the differences of $E_{Glob}$ between two age groups are significant. The results show higher global efficiency for age group +18. The average $E_{Glob}$ for high age group is 0.0685 while is 0.06 for low age group. These small values of $E_{Glob}$ imply that there are only very few vertices and edges of graphs or a small number of connections that exist in the brain. This is also known as *small-world* architecture of the brain network [40, 41].

VI. CONCLUSION

In this study, our goal was to assess differences between two age groups for the task of brain development analysis based on directional connectivity. Using Granger causality and reduced sine hyperbolic polynomial function, we were able to test for nonlinearity in brain connectivity. One of the most significant findings in this study was that causal interactions between brain regions revealed new connections between brain regions. To our knowledge, this study is among the first to use causal discovery in the context of brain development.

We found that three brain networks, i.e., DMN, VIS, and SAL, are causal activation of other brain regions. In addition, four brain regions DMN, SAL, AUD, and VNT are often led by other brain regions. The largest number of within region connections is in the DMN brain region. For between region connectivity, the two pairs (DMN, VIS) and (DMN, VENT) have most directional connections. Our findings with PNC data highlight the importance of causality modeling for the tasks of brain development analysis and brain age prediction.

The second goal of this study was to investigate the fusion of two types of brain connections, i.e., EC + FC, to assess how this fusion can improve the task of brain age prediction. We gained more than 4% of classification accuracy. We also found that the strength of GCI values exists in old age group. It indicates that stronger connections exist between brain ROIs as brain matures with age. The findings of this investigation with the fusion of FC and EC complement earlier studies based on FC only.

While the EC-based developmental brain connectivity analysis is promising, more studies are needed to determine and validate the causations. Using other imaging modalities may provide additional evidence and shed light on the use of causality in the study of brain maturation. It may also be worthwhile to examine conditional GC, in which the computation of granger causality value between two regions is calculated given the third brain region. This may also address the issues with the GC model [13, 16].